\documentclass[lettersize,journal]{IEEEtran}
\usepackage{amsmath,amsfonts}
\usepackage{algorithmic}
\usepackage{algorithm}
\usepackage{array}
\usepackage[caption=false,font=normalsize,labelfont=sf,textfont=sf]{subfig}
\usepackage{textcomp}
\usepackage{stfloats}
\usepackage{url}
\usepackage{verbatim}
\usepackage{graphicx}
\usepackage{cite}
\usepackage{enumitem}
\usepackage{booktabs}
\usepackage[hidelinks]{hyperref}
\begin{document}

\title{CABS+: Efficient and Scalable Model Merging via Conflict-Aware Sparsification and Adaptive Weight Allocation}



\author{Yuchen Liu, Zongzhen Yang, Binhang Qi, Hailong Sun*, Xiang Gao%
\thanks{Yuchen Liu, Zongzhen Yang, Binhang Qi, Hailong Sun, and Xiang Gao are with the State Key Laboratory of Complex \& Critical Software Environment (CCSE), Beihang University, Beijing, China, and the Hangzhou Innovation Institute of Beihang University, Hangzhou, China.}%
\thanks{Binhang Qi is also with the National University of Singapore, Singapore.}%
}


\maketitle

\begin{abstract}
Model merging has recently attracted significant attention as a promising  paradigm for constructing unified multi-task models without requiring additional retraining. 
However, due to the widespread presence of parameter conflicts and knowledge interference across tasks, the performance of merged models is often unsatisfactory.
To address these challenges, prior work introduced the Conflict-Aware and Balanced Sparsification (CABS) method, which reduces parameter interference through structured pruning and sequential masking. However, CABS, like the mainstream model merging methods, relies on grid search to determine scaling coefficients, leading to exponential time complexity, limiting its practical applicability. Meanwhile, its optimization objective is prone to being dominated by high performance tasks, resulting in suboptimal overall performance.

To overcome these limitations, we extend CABS and propose the enhanced method, termed CABS+. 
Specifically, the Adaptive Weight Allocation (AWA) strategy optimizes merging coefficients via a gradient-free search scheme to reduce time complexity, while the asymmetric fitness function promotes more comprehensive performance gains across tasks.
Moreover, to better understand the key factors influencing model merging performance, we conduct the systematic empirical study and propose a new metric, Relative Synergy Score (RSS), to quantify model mergeability, providing practical guidance for model selection in real-world applications.
We compare CABS+ with state-of-the-art model merging methods, including CABS, AdaMerging, and WUDIMerging, across 27 datasets and 5 models covering large language models, small-scale language models, and vision models.
Extensive experiments across diverse tasks and model scales verify the effectiveness and efficiency of CABS+.
Compared with AdaMerging and WUDIMerging, CABS+ achieves overall performance improvements of 16.97\% and 12.93\%, exhibits stronger stability and robustness under varying task numbers and model architectures, while requiring less than 25\% of the GPU memory used by AdaMerging and achieving nearly a 4$\times$ speedup in merging time compared with WUDIMerging.
\end{abstract}


\section{Introduction}
\IEEEPARstart{W}{ith} the widespread adoption of the pretraining finetuning paradigm, a large number of task-specific model checkpoints have been accumulated in the open-source community. However, directly deploying multiple specialized models for different tasks incurs substantial storage and maintenance overhead, which is particularly undesirable in resource-constrained or real-world application scenarios \cite{kim2025task}. Although multi-task learning can partially alleviate this issue, it typically relies on joint training over multiple tasks, resulting in considerable computational cost. Moreover, it often requires access to aggregated multi-task data, which may raise concerns regarding data sharing and privacy \cite{du2024parameter}. To address these limitations, model merging has emerged as an efficient alternative paradigm. The key idea is to combine multiple expert models in the parameter space to construct a unified multi-task model, without requiring access to multi-task training data or additional retraining.

In recent years, model merging has been extensively studied in the deep learning community and has demonstrated promising practical utility across various domains \cite{chen_toward_2026,li_deep_2025}. In the context of large language models (LLMs), task vector approaches have emerged as a dominant paradigm \cite{jin_dataless_2022,akiba_evolutionary_2025,ilharco2023editing,yang2024adamerging,cheng2025whoever}.
However, due to the widespread presence of parameter conflicts and knowledge interference among expert models, the merged model often exhibits inferior performance on individual tasks compared to its corresponding single-task counterparts \cite{zhang2025channel}.
To address this issue, recent studies have explored model merging from the perspective of parameter space geometry. In particular, sparsification of task vectors has been shown to effectively reduce parameter conflicts and has emerged as a strong paradigm for improving merging performance \cite{he2026mergebench,yu2024language,yadav_ties-merging_2023,davari2024model,he_localize-and-stitch_2024}.
Such sparsification-based methods
are closely related to model pruning techniques, where magnitude-based pruning \cite{liang_pruning_2021} is commonly used to retain important parameters while removing redundant updates. However, prior studies \cite{yu2024language} have observed that, in the context of model merging, magnitude-based pruning can be less effective than random sparsification. This is primarily because it tends to preserve highly overlapping and unevenly distributed parameters, which may exacerbate inter-task interference and limit the final performance.

To address the limitations of existing sparsification-based model merging methods, we previously proposed the Conflict-Aware and Balanced Sparsification method (CABS, accepted by ICML 2025). CABS performs cross-layer uniform pruning to maintain the balanced distribution of information within each task vector, and further incorporates a sequential masking mechanism to encourage separation among task vectors in the parameter space, thereby reducing potential conflicts. However, CABS relies on grid search to determine task scaling coefficients, in which time complexity grows exponentially with the number of tasks to be merged, leading to high practical overhead. In addition, its optimization objective focuses on the aggregated performance across all tasks, which tends to bias the resulting scaling coefficients toward a subset of high-performing tasks, thereby weakening performance on the remaining ones. Existing approaches for optimizing merging coefficients can be broadly categorized into two groups. The first introduces additional modules to model data features or meta-information, typically incurring extra training cost and computational overhead \cite{zhang2024knowledge,chen2026learn}. The second category determines scaling coefficients using test data without requiring additional training data, as exemplified by AdaMerging \cite{yang2024adamerging}. However, such methods often demand substantial GPU memory when applied to large-scale language models with billions of parameters and long-context inputs, with memory consumption growing linearly with the number of models to be merged, which makes them less feasible on widely accessible hardware platforms (e.g., V100 GPUs) and limits the scalability of model merging in practical deployments.

To address the aforementioned limitations, this work extends CABS and proposes an enhanced framework, termed CABS+.
Specifically, CABS+ introduces the \textbf{Adaptive Weight Allocation (AWA)} strategy, which performs efficient optimization via the gradient-free search mechanism. By eliminating the need for gradient computation and backpropagation, AWA reduces the memory overhead of the merging process to the level of standard inference, making it feasible to merge billion parameter models on a single GPU with limited memory. Meanwhile, AWA incorporates the boundary-constrained search space along with the asymmetric fitness evaluation strategy to account for scale differences across task-specific losses. Compared to conventional black-box optimization methods, this design facilitates faster convergence and mitigates the tendency of the optimization process to be dominated by high-loss tasks. In addition, AWA is naturally compatible with the CABS pruning strategy. Since task vectors have already undergone conflict reduction during the CABS pruning stage, the resulting optimization landscape for merging coefficients becomes smoother with reduced non-convexity. This property provides favorable conditions for efficient search and stable convergence of AWA, thereby improving the efficiency and reliability of the optimization process. 
Furthermore, to fill the gap in the community regarding the lack of systematic exploration into key factors affecting model merging, we conduct the multi-dimensional empirical study on model mergeability and propose \textbf{Relative Synergy Score (RSS)} to quantify it. Through extensive analysis, multiple key factors influencing mergeability are identified and validated, including task heterogeneity, data distribution heterogeneity, training configurations, model architecture, and model scale, thereby providing practical guidelines for model selection prior to merging.


Extensive experimental results show that CABS+ achieves superior performance across various settings while also offering notable efficiency advantages.
In terms of performance, CABS+ demonstrates stronger robustness to variations in model architectures and the number of tasks, and achieves higher average performance than state-of-the-art methods, outperforming AdaMerging and WUDIMerging by 16.97\% and 12.93\%, respectively. 
In representative large-model efficiency experiments on Mistral, CABS+ requires less than 25\% of the GPU memory used by AdaMerging and achieves a nearly 4$\times$ speedup in merging time compared with WUDIMerging.

The main contributions of this work are summarized as follows:
\begin{itemize}
\item{We propose CABS+, which extends CABS to address its high time complexity, reduces excessive GPU memory consumption of comparable methods, and mitigates the tendency of optimization to be dominated by tasks with larger loss scales, thereby achieving more comprehensive performance improvements across tasks.}
\item{We conduct the systematic investigation of model mergeability and identify six critical factors that affect model merging performance. To support this analysis, we propose the new metric, termed Relative Synergy Score, to quantify model mergeability, thereby providing practical guidelines for model selection before merging.}
\item{Extensive experiments across 27 tasks and 5 models of varying scales demonstrate that CABS+ outperforms AdaMerging and WUDIMerging by 16.97\% and 12.93\% in overall performance, respectively, and exhibits advantages in both memory usage and merging efficiency.}
\end{itemize}
Our source code is available at \url{https://anonymous.4open.science/r/CABS_Plus-70C1}.

\section{Related Work}

\subsection{Data-Free Model Merging}
The most basic merging strategy is simple averaging \cite{izmailov2018averaging,wortsman_model_2022}, which directly computes the mean of corresponding parameters across models. However, this approach often fails to account for task-specific parameter variations, leading to suboptimal performance. To achieve more refined weight allocation, Fisher Merging \cite{matena_merging_2022} introduces the Fisher information matrix to assess the importance of each expert model’s parameters and assign corresponding merging weights accordingly. Similarly, RegMean \cite{jin_dataless_2022} adopts an alternative approach by minimizing the prediction discrepancy between the merged model and individual task models to merge. Subsequently, Task Arithmetic \cite{ilharco2023editing} proposes an innovative method based on task vectors, defined as the difference between the finetuned model and the pretrained model parameters, and employs scaling coefficients to flexibly control the contribution of each task vector during the merging process. WUDI-Merging \cite{cheng2025whoever} considers the geometric relationships of task vectors in parameter space, characterizing the linear subspace structure and suppressing inter-task interference through the optimization process. Additional approaches focus on aligning the losses between the merged model and the task-specific models \cite{xiong2024multi,pmlr-v267-wei25k}. Although such methods can effectively merge models, significant parameter redundancy and sign conflicts often exist between expert models, which can severely degrade the performance of the merged model. 
\subsection{Test-Time Adaptive Model Merging}
Test-Time adaptive methods employs certain test data to resolve conflicts between tasks. For instance, AdaMerging \cite{yang2024adamerging} minimizes the unsupervised entropy on test samples as an objective and dynamically learns the merging coefficients via gradient descent. Representation Surgery \cite{wei2025representation} seeks to minimize the discrepancy between the merged model and individual expert models in feature space, thereby effectively alleviating representation bias during feature extraction. However, such methods frequently incur substantial GPU memory usage when applied to large scale models, thereby limiting their feasibility in resource-constrained environments.
\subsection{Sparsification-Based Model Merging}
Sparsification-based methods primarily target the removal of redundant or conflict parameters through sparsification. TIES-Merging \cite{yadav_ties-merging_2023} preserves critical knowledge by pruning low-magnitude redundant parameters and resolving sign conflicts. Consensus Merging \cite{wang2024localizing} further enhances performance by eliminating weights that negatively impact the merging process, often referred to as selfish or catastrophic weights. DARE \cite{yu2024language}, inspired by the Dropout \cite{srivastava2014dropout} mechanism, randomly discards a large portion of parameters and rescales the remaining weights, demonstrating the potential of sparsity to alleviate conflicts. These sparsification strategies, which have shown remarkable effectiveness in model merging, are closely related to conventional model compression and pruning techniques used to reduce computational cost while retaining core performance. Notably, magnitude-based pruning assumes that parameters with larger absolute values carry more critical information \cite{kovaleva_bert_2021,puccetti_outlier_2022,yin_outlier_2024}. However, in the context of model merging, directly applying magnitude pruning can lead to highly uneven weight distributions, exacerbating inter-task conflicts rather than mitigating them.

To address this imbalance in weight distribution, structured pruning techniques were incorporated into the model merging process. The Conflict-Aware and Balanced Sparsification method (CABS, accepted by ICML 2025), employs sequential pruning to generate masks that eliminate parameter overlap among task vectors, while integrating $n:m$ pruning \cite{zhou2021learning,xia_structured_2022} to ensure globally balanced weight distributions. This approach significantly improves both the stability and performance of multi-task model merging. However, CABS relies on the grid search strategy to determine the scaling coefficients, resulting in prohibitively high time complexity and limiting its practical applicability. Moreover, its optimization objective typically aggregates the performance of all tasks in a simple manner, which can lead to uneven performance improvements across tasks and suboptimal overall outcomes.

\section{Methodology}

\begin{figure*}[!t]
\centering
\includegraphics[width=\textwidth]{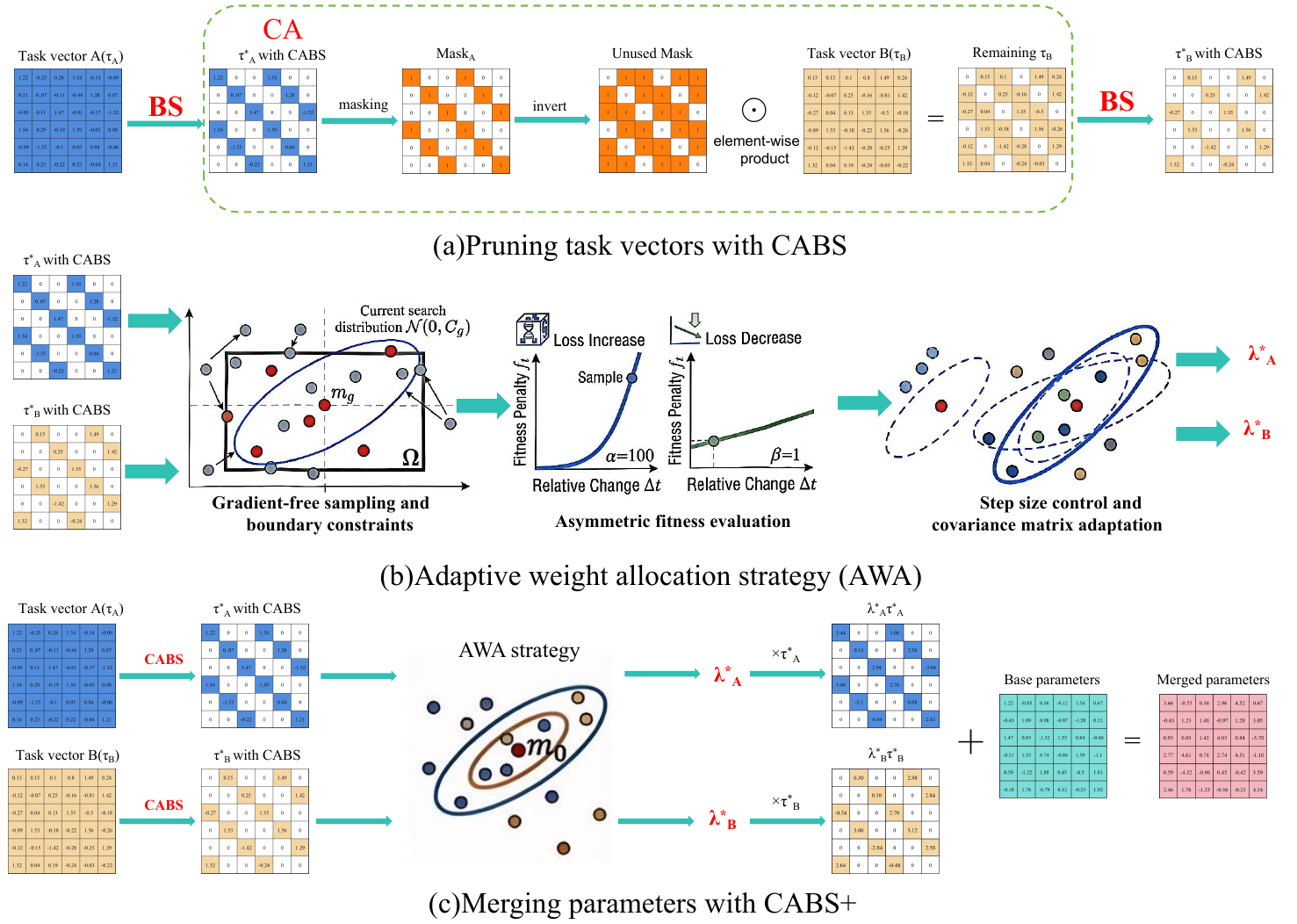}
\caption{Illustration of the overall framework of CABS+. (a) Conflict-Aware and Balanced Sparsification; (b) Adaptive Weight Allocation strategy; (c) Overall pipeline of CABS+.}
\label{framework}
\end{figure*}

To address the above issues, we propose CABS+, whose overall pipeline is illustrated in Fig. \ref{framework}. 
Compared with CABS, CABS+ mainly differs in the introduction of a gradient-free adaptive weight allocation (AWA) strategy, which reduces the time complexity of the previous grid-search approach while improving model merging performance. The detailed procedure is given in Algorithm \ref{alg:cabs_awa}.
\subsection{Conflict-Aware Sparsification (CA)}
\textbf{Sequential pruning and masking.} CA mainly adopts a sequential pruning strategy to avoid parameter overlap between task vectors, thereby effectively eliminating parameter conflicts during merging. Specifically, task vector $\tau_A$ is first pruned, and the positions of the parameters to be retained are marked to generate a mask, denoted as ${mask}_A$. Subsequently, this mask is used to guide the pruning of the subsequent task vector $\tau_B$, so as to prevent parameter overlap at the same positions. More specifically, in order to eliminate the overlap with $\tau_A$, the parameters retained in task vector $\tau_B$ are computed as follows before the subsequent pruning:
\begin{equation} 
\tau_\text{B remaining}=\tau_B\odot(1-\mathrm{mask}_A).
\end{equation}
Afterward, the task vector $\tau_\text{B remaining}$ is pruned to generate the corresponding new ${mask}_B$. Then, the pruned task vector $\tilde{\tau}_B$ is merged with the previously pruned task vector $\tilde{\tau}_A$ to obtain a model parameter matrix without overlap. If multiple tasks exist, the same procedure is applied.

\textbf{Minimizing overlap under low sparsity.} When the sum of the retained parameter ratios of all task vectors exceeds 1, overlap cannot be completely avoided. For example, if both task A and task B retain 60\% of the parameters after pruning, only 40\% of the parameter space in task vector $\tau_B$ can remain fully non-overlapping with $\tau_A$. The remaining 20\% parameters of $\tau_B$ are therefore selected from the overlapping region. For this unavoidable overlap, sign selection and parameter averaging strategies similar to TIES-Merging \cite{yadav_ties-merging_2023} are adopted to preserve the performance of both tasks as much as possible. This overlap-restricted strategy consistently provides stable performance improvements across different tasks and model settings. 
A detailed derivation showing that the non-overlapping masks generated by CA make the pruned task vectors orthogonal in the Frobenius inner product, eliminate the cross term in the merged update norm, and enable independent scaling of task-vector contributions is provided in the supplementary material.

\begin{algorithm}[t]
\caption{CABS+}
\label{alg:cabs_awa}

\small
\begin{tabular}{@{}l p{0.78\linewidth}@{}}
\textbf{Input:} & Task vectors $\tau_A, \tau_B$, base model $W_{\text{base}}$, sparsity level $n, m$, feasible bounds $[l, u]$, population size $K$, max generations $G$ \\
\textbf{Output:} & Merged model parameters $W_{\text{final}}$
\end{tabular}
\vspace{0.2em}
\begin{algorithmic}[1]
\STATE \textbf{Phase 1: Conflict-Aware and Balanced Sparsification (CABS)}
\vspace{-1em}
\STATE Apply $n:m$ pruning to $\tau_A$ and compute $\text{mask}_A$ 
\STATE $\tau_{B \text{ remaining}} = \tau_B \odot (1 - \text{mask}_A)$ to eliminate overlap with $\tau_A$ 
\STATE Apply $n:m$ pruning to $\tau_{B \text{ remaining}}$ to compute $\text{mask}_B$ 
\STATE Define pruned vectors: $\tilde{\tau}_A = \text{mask}_A \odot \tau_A$, $\tilde{\tau}_B = \text{mask}_B \odot \tau_B$
\STATE \textbf{Phase 2: Adaptive Weight Allocation (AWA)}
\vspace{0.2em}
\STATE Initialize: covariance $C^{(0)} = I$, step size $\sigma^{(0)} = 0.05$
\STATE Compute base loss $L_{\text{base}, t}$ for each task $t$ at initial state $\lambda_0 = \mathbf{1}$
\FOR{$g = 0$ to $G-1$}
    \FOR{$k = 1$ to $K$}
        \STATE Sample coefficients: $\lambda_k^{(g)} \sim m^{(g)} + \sigma^{(g)} \mathcal{N}(0, C^{(g)})$
        \STATE Bound constraint $P_{\Omega}$: $\lambda_{k,i}^{(g)} = \min(u, \max(l, \lambda_{k,i}^{(g)}))$
        \STATE \parbox[t]{0.90\linewidth}{\raggedright Construct candidate model:
        $\begin{aligned}
        W_k = W_{\text{base}}
        &+ \lambda_{k,A}^{(g)} \tilde{\tau}_A + \lambda_{k,B}^{(g)} \tilde{\tau}_B
        \end{aligned}$}
        \STATE Evaluate asymmetric fitness: $F(\lambda_k^{(g)}) = \sum_{t} f_t(\lambda_k^{(g)})$
    \ENDFOR
    \STATE Sort candidates by $F(\lambda_k^{(g)})$ and select top $\mu = K/2$ optimal samples
    \STATE Update $m^{(g+1)}$ using logarithmically weighted average of top $\mu$ samples
    \STATE Update $\sigma^{(g+1)}$ and $C^{(g+1)}$ using evolution paths $p_\sigma$ and $p_c$
\ENDFOR
\STATE Extract optimal coefficients: $\lambda^* = \arg\min F(\lambda)$
\STATE Merge the pruned vectors with the base model using optimal coefficients:

\quad $W_{\text{final}} = W_{\text{base}} + \lambda_A^* \times \tilde{\tau}_A + \lambda_B^* \times \tilde{\tau}_B$
\STATE Return $W_{\text{final}}$
\end{algorithmic}
\end{algorithm}

\subsection{Balanced Sparsification (BS)}
The weight matrix of the model is first divided into m non-overlapping blocks of consecutive weights. Within each block, only the n parameters with the largest absolute values are retained, and the rest are pruned. This block-wise local pruning strategy is applied uniformly across all layers of the model, so that the retained weights are distributed more evenly throughout the network rather than concentrated in one or two layers, thereby preventing severe parameter conflicts when multiple task models are merged.


It is worth noting that the proposed BS strategy differs fundamentally from traditional n:m pruning in terms of its objective. Conventional n:m pruning primarily aims to reduce computational and memory costs through structured sparsity for model compression and inference acceleration. In contrast, BS is designed to mitigate parameter conflicts during model merging by enforcing more balanced feature distributions across different task vectors with higher sparsity ratios. Consequently, unlike conventional sparse models, the final merged model produced by BS remains dense. This strategy prioritizes merging performance and task compatibility rather than inference efficiency.

\subsection{Adaptive Weight Allocation (AWA)}
CA effectively ensures orthogonality among task vectors in the parameter space to minimize conflicts, while BS guarantees the uniform retention of each task vector’s information to enhance stability. However, achieving better model merging performance still faces a key challenge: How to determine the scaling coefficient for each task vector? Even if the task vectors are already orthogonal and evenly distributed in the parameter space, the scaling coefficients $\lambda$ that determine their contributions must still be set when merging them into the base model. Improper allocation of these coefficients can significantly degrade the performance of the merged model.

In the previous conference version, we used grid search to determine these coefficients. However, this approach has two critical limitations. First, its computational time complexity grows exponentially with the number of tasks being merged. Second, the objective focuses solely on the sum of performance across all tasks. This often causes the selected coefficients to be dominated by a few high-performing tasks and prevents achieving optimal merging performance. Existing methods for optimizing merging coefficients, such as AdaMerging, mostly rely on gradient-based optimization. This requires constructing the full computation graph during the forward pass to enable backpropagation through the chain rule. In the context of model merging, multiple task vectors must be simultaneously incorporated into the computation graph, as they jointly participate in the optimization of merging coefficients. This requires the GPU to maintain not only the base model parameters but also all task-specific parameter updates at the same time. To compute loss.backward(), the intermediate results of each layer must be stored. For large language models with billions of parameters and contexts of several thousand tokens, storing the intermediate activations needed for gradient computation consumes enormous GPU memory. When multiple task vectors are involved, this memory overhead is further amplified, as each task contributes additional parameters and corresponding computation paths in the graph. Specifically, memory complexity grows linearly with the number of task vectors, as well as model depth $L$ and sequence length $T$, leading to substantial memory overhead in model merging scenarios. Even with a batch size of 1, this often exceeds the memory capacity of a single high-performance GPU such as the A100. This memory bottleneck severely limits the scalability of model merging techniques.

To address these issues, we propose a new computationally efficient adaptive weight allocation strategy, which is a gradient-free automatic optimization mechanism improved based on the Covariance Matrix Adaptation Evolution Strategy (CMA-ES) \cite{hansen2001completely}. Since this method does not require gradient computation or backpropagation, the GPU memory usage is strictly limited to the level of the inference stage, making it possible to merge billion parameter scale models on consumer-level GPUs such as the V100. At the same time, this strategy can be well combined with the CA and BS methods described above. In general, evolutionary algorithms tend to suffer from slow convergence. However, because the task vectors can minimize task conflicts after the CABS pruning process, the optimization landscape of the merging coefficients becomes smoother and the non-convexity is significantly reduced. This allows the proposed AWA optimization strategy to perform the search more efficiently and achieve faster convergence, thereby improving the stability and reliability of the overall optimization process.

\textbf{Gradient-free sampling and boundary constraints.}  AWA formulates the determination of the scaling coefficients as an optimization problem in a continuous space. Specifically, $K$ candidate solutions are first sampled from a multivariate normal distribution:
\begin{equation}
\lambda_k^{(g)}\sim\mathbf{m}^{(g)}+\sigma^{(g)}\mathcal{N}(\mathbf{0},\mathbf{C}^{(g)}),\quad k=1,...,K.
\end{equation}
Here, $g$ denotes the iteration index, ${m}^{(g)}$ is the mean vector of the current distribution, and $C$ is the covariance matrix of the algorithm with the initial value set to the identity matrix $I$. The step size is denoted by $\sigma$, whose initial value is set to 0.05 by default. The coefficient vector is denoted by $\lambda$, whose dimension $D$ equals the number of tasks to be merged.

To ensure that the scaling coefficients remain within a reasonable range rather than exploring an unbounded space, we introduce boundary constraints based on the CMA-ES algorithm. The feasible region $\Omega = \{\mathbf{v} \in \mathbb{R}^D \mid l \le v_i \le u,\ \forall i = 1,\dots,D\}$ is defined in a $D$ dimensional space, where the lower bound is set to $l$=0.1 and the upper bound is set to $u$=2. At the $g$-th iteration, for each $\lambda_k^{(g)}$, the projection function ${P}_\Omega$ is applied to obtain the constrained coefficient $\lambda_k^{(g)}$ :
\begin{equation}
\lambda_k^{(g)}=\mathcal{P}_\Omega\left(\mathbf{m}^{(g)}+\sigma^{(g)}\mathcal{N}(\mathbf{0},\mathbf{C}^{(g)})\right).
\end{equation}
The explicit expression in component-wise form is given as:
\begin{equation}
\left[\lambda_k^{(g)}\right]_i=\min\left(u,\max\left(l,\left[\lambda_k^{(g)}\right]_i\right)\right),\quad i=1,...,D.
\end{equation}

\textbf{Asymmetric fitness evaluation for handling scale differences.} In the standard CMA-ES strategy, the optimization objective is to minimize the sum of the absolute loss values over all tasks:
\begin{equation}
J_{\mathrm{trad}}(\lambda)=\sum_{t=1}^TL_t\left(\theta(\lambda)\right).
\end{equation}
where $\theta$ denotes the merged model. However, since different tasks have different loss scales, the optimizer tends to be dominated by tasks with larger losses while ignoring tasks with smaller losses. According to our experimental observations, this typically manifests as improving the performance of some tasks at the cost of degrading the performance of others during the optimization process.

To address this question, we introduce the asymmetric penalty mechanism in AWA. For each individual $\lambda_k^{(g)}$, the fitness function ${F}\left(\lambda_k^{(g)}\right)$ is computed. Specifically, before the search begins, the baseline loss of each task under the initial state $\lambda_k^{(g)}$ is first computed:
\begin{equation}
L_{\mathrm{base}}^{(t)}=\mathcal{L}_t\left(\mathbf{\theta}(\lambda^{(0)})\right).
\end{equation}
For any candidate solution $\lambda$, the normalized relative change with respect to the baseline is computed to eliminate scale differences across tasks:
\begin{equation}
\Delta_t(\lambda)=\frac{L_t(\mathbf{\theta}(\lambda))-L_{\mathrm{base}}^{(t)}}{L_{\mathrm{base}}^{(t)}}.
\end{equation}
Subsequently, the asymmetric penalty function $f_t$ is constructed, which applies a large penalty to tasks whose loss has increased, meaning that they have experienced performance drop, for example with $\alpha$=100 and $\beta$=1.
\begin{equation}
f_t(\lambda)=
\begin{cases}
\alpha\cdot\Delta_t(\lambda) & \mathrm{if}\Delta_t(\lambda)>0. \\
\beta\cdot\Delta_t(\lambda) & \mathrm{if}\Delta_t(\lambda)\leq0.
\end{cases}
\end{equation}
The final fitness function is obtained as the sum of the asymmetric scores across all tasks:
\begin{equation}
F(\lambda)=\sum_{t=1}^Tf_t(\lambda)
\end{equation}

\textbf{Population selection and mean update.} The samples are ranked according to $F(\lambda)$, and the top $\mu$ samples ($\mu=K/2$) are selected to update the distribution parameters ${m}^{(g+1)}$, $\sigma^{(g+1)}$, ${C}^{(g+1)}$ in order to maximize the likelihood estimate. The new distribution mean 
${m}^{(g+1)}$ is computed as the weighted average of the top $\mu$ individuals. Let 
$\lambda_{1:K}^{(g)},...,\lambda_{\mu:K}^{(g)}$ denote the top $\mu$ individuals in the current iteration, then:
\begin{equation}
\mathbf{m}^{(g+1)}=\sum_{i=1}^\mu w_i\boldsymbol{\lambda}_{i:K}^{(g)}.
\end{equation}
where $w_i$ are the normalized weights ($w_1>\cdots>w_\mu>0\text{,}\Sigma w_i=1$), representing the shift of the search center. To ensure that higher-ranked individuals contribute more to the mean update, the weights are typically defined as:
\begin{equation}
w_i^{\prime}=\ln(\mu+0.5)-\ln(i),\quad\mathrm{for~}i=1,...,\mu.
\end{equation}
They are then normalized so that their sum equals 1:
\begin{equation}
w_i=\frac{w_i^{^{\prime}}}{\sum_{j=1}^\mu w_j^{^{\prime}}}.
\end{equation}
where $w_1$ is the largest and exerts the strongest influence in pulling the mean $m$ toward its position,
while $w_\mu$ is the smallest, and all weights less than $w_\mu$ are set to zero, contributing no further updates.

\textbf{Step size control and covariance matrix adaptation.} AWA uses the conjugate evolution path to control the step size $\sigma$. The principle is as follows: if the evolution path remains approximately orthogonal over multiple iterations, indicating a random walk, the step size is too small; if the path direction remains consistent over iterations, indicating straight-line movement, the step size is too large.

First, the vector ${p}_\sigma$ that records the step size path is updated:
\begin{equation}
\begin{split}
\mathbf{p}_\sigma^{(g+1)} &= (1-c_\sigma)\mathbf{p}_\sigma^{(g)} \\
&\quad +\sqrt{c_\sigma(2-c_\sigma)\mu_{\mathrm{eff}}}\cdot\left(\mathbf{C}^{(g)}\right)^{-\frac12}
\frac{\mathbf{m}^{(g+1)}-\mathbf{m}^{(g)}}{\sigma^{(g)}}.
\end{split}
\end{equation}
where $c_\sigma$ is the decay time constant, ${C}^{-\frac12}$ denotes the isotropic transformation of the covariance matrix, and $\mu_{\text{eff}}$ is the variance-effective selection mass:
\begin{equation}
\mu_{\mathrm{eff}}=\frac{1}{\sum_{i=1}^\mu w_i^2}=\frac{1}{\|\mathbf{w}\|^2}.
\end{equation}
The step size $\sigma^{(g+1)}$ is then updated based on ${p}_\sigma^{(g+1)}$:
\begin{equation}
\sigma^{(g+1)}=\sigma^{(g)}\exp\left(\frac{c_\sigma}{d_\sigma}\left(\frac{\parallel\mathbf{p}_\sigma^{(g+1)}\parallel}{E\parallel\mathcal{N}(\mathbf{0},\mathbf{I})\parallel}-1\right)\right).
\end{equation}
At this stage, if the path length $\parallel{p}_\sigma\parallel$ exceeds the expected length for a random walk, $\sigma$ is increased; otherwise, it is decreased. Simultaneously, the evolution path ${p}_c$, which records the anisotropy of the search trajectory, is updated:
\begin{equation}
\mathbf{p}_c^{(g+1)}=(1-c_c)\mathbf{p}_c^{(g)}+\sqrt{c_c(2-c_c)\mu_{\mathrm{eff}}}\frac{\mathbf{m}^{(g+1)}-\mathbf{m}^{(g)}}{\sigma^{(g)}}.
\end{equation}
Finally, the covariance matrix is updated using the decayed previous matrix $c^{(g)}$, the historical evolution path ${p}_c^{(g+1)}$, and the information from the current population:
\begin{equation}
\begin{split}
\mathbf{C}^{(g+1)} =& (1-c_1-c_\mu)\mathbf{C}^{(g)} \\
&+ c_1\left(\mathbf{p}_c^{(g+1)}\left(\mathbf{p}_c^{(g+1)}\right)^T\right) \\
&+ c_\mu\sum_{i=1}^\mu w_i\,\mathbf{y}_i\mathbf{y}_i^T.
\end{split}
\end{equation}
where $\mathbf{y}_{i}={(\lambda_{i:K}^{(g)}-\mathbf{m}^{(g)})}/{\sigma^{(g)}}$ denotes the deviation vectors after subtracting the mean.

The adaptive update of the covariance matrix $C$ endows AWA with a fundamental advantage beyond simple heuristic search. It can not only dynamically adjust the scale of the search space, but also implicitly learn the relationships between different task coefficients. For example, if the optimization landscape exhibits off-diagonal characteristics, such as an increase in $\lambda_1$ requiring a specific proportional decrease in $\lambda_2$ to maintain loss reduction, AWA automatically adapts the search distribution to align with the objective function’s contours, forming a hyperellipsoid that allows efficient descent along correlated directions. This optimization is unattainable with grid search, which evaluates each task coefficient independently.

\textbf{Final output and model merging.} After the search finishes, the optimal coefficient vector is obtained:
\begin{equation}
\lambda^*=\mathrm{argmin}F(\lambda).
\end{equation}
The final model merging is then performed:
\begin{equation}
\theta_{merged}=\theta_{base}+\sum_{t=1}^T\lambda_t^*\cdot\tilde{\tau}_t.
\end{equation}
where $T$ denotes the number of models to be merged, and $\tilde{\tau}_t$ represents the task vectors processed by CABS pruning.

\textbf{Time complexity comparison.} Compared with grid search, AWA has the significant advantage in time complexity. Assume there are $T$ tasks, and $S$ search steps are sampled in each dimension. Grid search requires $\mathcal{O}(S^T)$ evaluations, which makes it prone to the curse of dimensionality when merging multiple tasks, resulting in a substantial increase in time consumption. In contrast, the complexity of AWA mainly depends on the population size $K$ and the number of iterations $G$. The upper bound of the total number of evaluations is $\mathcal{O}(K \times G)$, which avoids the exponential dependence on the number of tasks. This not only significantly reduces the time required for merging, but also enables the search for better solutions in the wider continuous space.

\section{Experiments}

\subsection{Experimental Setup}

\textbf{Benchmarks and Checkpoints.}
We conduct large language model evaluations on the LLM Leaderboard benchmark \cite{open-llm-leaderboard} using the Mistral-7B-v0.1 \cite{jiang_mistral_2023} backbone and its fine-tuned variants WildMarcoroni-Variant1-7B and WestSeverus-7B-DPO-v2. This benchmark includes six tasks: AI2 Reasoning Challenge \cite{clark2018think}, HellaSwag \cite{zellers2019hellaswag}, MMLU \cite{wang_mmlu-pro_2024}, TruthfulQA \cite{lin_truthfulqa_2022}, Winogrande \cite{sakaguchi2021winogrande}, and GSM8K \cite{cobbe2021training}. In addition, we perform experiments on the Open LLM Leaderboard 2 \cite{open-llm-leaderboard-v2} benchmark using Qwen-2.5-7B-Instruct \cite{yang_qwen2_2024-1} and its fine-tuned variants fq2.5-7B and Tsunami-0.5-7B. This benchmark consists of six more challenging tasks, including IFEval \cite{zhou2023instructionfollowingevaluationlargelanguage}, BBH \cite{suzgun2022challengingbigbenchtaskschainofthought}, MATH \cite{hendrycks2021measuringmathematicalproblemsolving}, GPQA \cite{rein2023gpqagraduatelevelgoogleproofqa}, MUSR \cite{sprague2024musrtestinglimitschainofthought}, and MMLU-Pro \cite{wang_mmlu-pro_2024}. All corresponding checkpoints are publicly available on Hugging Face. The evaluations on both benchmarks are conducted using the EleutherAI Language Model Evaluation Harness \cite{eval-harness}, which is a widely recognized standard framework for evaluating LLM capabilities.

For smaller models, we evaluate RoBERTa \cite{liu2019roberta} and GPT-2 \cite{radford2019language} on the GLUE benchmark \cite{wang2018glue}. Specifically, we consider the CoLA \cite{warstadt2019neural}, MNLI\cite{nangia2017repeval}, MRPC \cite{dolan2005automatically}, QNLI, QQP, RTE \cite{dagan2005pascal,bar2006second,giampiccolo2007third,bentivogli2009fifth}, and SST-2 \cite{socher2013recursive} datasets. To further increase task difficulty and diversity, we additionally include the multiple-choice reading comprehension task RACE \cite{lai2017race} and the question answering task SQuAD \cite{rajpurkar2016squad} in the RoBERTa experiments. These datasets and their corresponding finetuned checkpoints are obtained from the FusionBench \cite{tang_fusionbench_2025} repository, which is a comprehensive benchmark and unified library for model merging. Detailed descriptions of all datasets and links to the corresponding checkpoints can be found in Sections B.7 and B.8 of the appendix in our conference version. 

\textbf{Evaluation Metrics.}
For tasks from the GLUE benchmark, we uniformly use accuracy as the evaluation metric. For tasks from the LLM Leaderboard benchmark and the Open LLM Leaderboard 2 benchmark, we adopt the default metrics specified in the corresponding leaderboard descriptions for each task, such as success rate and accuracy. Detailed descriptions of the metrics used can be found in Section B.9 of the appendix in our conference version.

\textbf{Baselines.}
We compare our CABS+ method with several representative and widely used model merging approaches, including Task Arithmetic and TIES-Merging \cite{yadav_ties-merging_2023}, as well as AdaMerging  \cite{yang2024adamerging}, which similar to our method, determines task scaling coefficients without requiring additional training data or architectural modifications. We also include our previous CABS method, which has been accepted by ICML 2025, as well as the recent state-of-the-art method WUDI-Merging \cite{cheng2025whoever}. In addition, since CABS+ involves a prepruning stage, we further combine the above methods with magnitude pruning and DARE pruning to evaluate their performance after pruning, thereby enabling a fairer comparison. 

To assess how far current model merging methods are from the expected ideal performance in large models, a new concept termed the “ideal model” is proposed. In our experiments, it is defined as the best performance achieved for each task among multiple checkpoints derived from the same base model under different fine-tuning configurations, serving as an upper bound for model merging performance.

\textbf{Implementation Details.} 
All model merging experiments are conducted on V100 GPUs with 32 GB memory on the Crater server platform \cite{noauthor_raids-labcrater_2026}. To ensure the consistency and stability of the results, each experimental configuration is evaluated three times, and the average results are reported. For large models, inference is performed using the lm-evaluation-harness version 0.4.0 library, with the batch size set to auto and all other parameters kept at their default values.

For the AWA strategy in CABS+, the population size is set to $K$=6, the initial step size is set to 
$\sigma$=0.05, and the number of iterations is set to $G$=50. An early stopping mechanism is applied such that the search terminates when the loss shows no significant change for 6 consecutive iterations. The remaining parameters follow the standard settings of the Covariance Matrix Adaptation Evolution Strategy. For the BS pruning strategy, the hyperparameters are consistent with those in the conference version. Specifically, the sparsity level is set to 0.90 for small models and 0.75 for large models.

\begin{table*}[!t]
\centering
\caption{Results of 7B LLMs on the Open LLM Leaderboard 2.}
\begin{tabular}{l|c|c|c|c|c|c|c}
\toprule
\textbf{Method} & \textbf{MMLU} & \textbf{IFEval} & \textbf{BBH} & \textbf{MATH} & \textbf{GPQA} & \textbf{MUSR} & \textbf{AVG} \\
\midrule
Tsunami-0.5-7b & 45.08 & 55.85 & 55.55 & 33.16 & 31.84 & 44.35 & 44.31 \\
fq2.5-7b & 44.83 & 44.97 & 56.23 & 34.36 & 32.13 & 46.57 & 43.18 \\
Ideal Model & 45.08 & 55.85 & 56.23 & 34.36 & 32.13 & 46.57 & 45.04 \\
\midrule
Task-Arithmetic & 44.83 & 43.75 & \textbf{56.17} & 28.36 & 31.31 & 46.72 & 41.86 \\
+ Magnitude & 44.91 & 55.14 & 55.36 & 35.01 & 31.65 & 43.28 & 44.23 \\
+ DARE & 44.71 & 45.49 & 55.71 & \underline{35.53} & 32.56 & 42.76 & 42.79 \\
TIES-Merging & \textbf{45.10} & 53.72 & 55.77 & 34.67 & 31.94 & 45.01 & 44.37 \\
+ DARE & 44.88 & 55.91 & 55.58 & 35.43 & 31.77 & 44.21 & 44.63 \\
AdaMerging & 44.74 & 47.77 & 55.99 & 31.37 & 31.46 & \underline{46.86} & 43.03 \\
+ Magnitude & 44.95 & 48.83 & 56.09 & 32.36 & 31.86 & 46.19 & 43.38 \\
+ DARE & 44.93 & 54.41 & 55.95 & 33.20 & 32.47 & 44.21 & 44.20 \\
WUDIMerging & \underline{45.05} & 53.54 & 55.96 & 33.18 & 31.79 & 45.00 & 44.09 \\
+ Magnitude & 44.99 & \underline{56.52} & 56.11 & 33.15 & 31.96 & 45.53 & 44.71 \\
+ DARE & 44.91 & 54.65 & 55.73 & 34.67 & \underline{32.72} & 43.16 & 44.31 \\
\midrule
CABS-fqFirst & 44.81 & 44.97 & 56.03 & 31.66 & 31.58 & \textbf{47.25} & 42.72 \\
\textbf{CABS+-fqFirst (ours)} & 44.63 & 55.82 & 56.03 & \textbf{35.73} & 31.99 & 46.31 & \underline{45.09} (+2.37) \\
CABS-TFirst & 44.86 & 44.45 & 55.92 & 30.37 & 31.66 & 46.19 & 42.24 \\
\textbf{CABS+-TFirst (ours)} & 44.55 & \textbf{57.32} & \underline{56.15} & 34.69 & \textbf{32.92} & 44.99 & \textbf{45.10} (+2.86) \\
\bottomrule
\end{tabular}
\label{tab:qwen2.5_merging_comparison}
\end{table*}

\begin{table*}[!t]
\centering
\caption{Performance comparison of 7B LLMs on LLM Leaderboard using different merging methods.}
\begin{tabular}{l|c|c|c|c|c|c|c}
\toprule
\textbf{Method} & \textbf{ARC} & \textbf{Hella.} & \textbf{MMLU} & \textbf{TQA} & \textbf{Wino.} & \textbf{GSM8K} & \textbf{AVG} \\
\midrule
WestSeverus & 71.30 & 88.26 & 63.92 & 72.72 & 83.69 & 74.27 & 75.69 \\
WildMarcoroni & 73.63 & 88.67 & 63.96 & 70.07 & 84.34 & 74.48 & 75.86 \\
Ideal Model & 73.63 & 88.67 & 63.96 & 72.72 & 84.34 & 74.48 & 76.30 \\
\midrule
Task Arithmetic & 72.52 & \underline{89.25} & 63.39 & \underline{74.00} & 83.46 & 73.38 & 76.02 \\
+ Magnitude & 71.93 & \textbf{89.32} & 63.18 & 73.85 & 84.12 & 72.22 & 75.77 \\
+ DARE & 72.64 & 88.86 & 63.54 & 72.82 & 84.03 & 73.44 & 75.89 \\
TIES-Merging & 71.42 & 89.17 & 63.16 & 73.82 & \textbf{84.74} & 73.01 & 75.89 \\
+ DARE & 71.87 & 88.95 & 63.56 & 72.87 & 84.61 & 73.21 & 75.85 \\
AdaMerging & 72.84 & 88.65 & 63.35 & 72.62 & 84.14 & 73.81 & 75.90 \\
+ Magnitude & 72.92 & 87.77 & 63.42 & 73.01 & 83.90 & 75.24 & 76.04 \\
+ DARE & 72.05 & 87.26 & 63.79 & 72.64 & 83.71 & \textbf{75.73} & 75.86 \\
WUDIMerging & \textbf{75.33} & 88.43 & 63.84 & 73.96 & 83.76 & \underline{75.62} & \textbf{76.82} \\
+ Magnitude & 74.05 & 87.04 & \textbf{63.97} & 64.68 & 80.13 & 75.55 & 74.24 \\
+ DARE & 73.70 & 87.97 & \underline{63.93} & 69.16 & 79.29 & 71.54 & 74.27 \\
\midrule
CABS & 72.92 & 88.89 & 63.50 & \textbf{74.41} & \underline{84.63} & 74.65 & 76.50 \\
\textbf{CABS+ (ours)} & \underline{74.20} & 89.15 & 63.67 & 73.76 & 84.52 & 74.93 & \underline{76.71} \\
\bottomrule
\end{tabular}
\label{tab:Mistral_results}
\end{table*}

\subsection{Results on Large Language Models}
Tables \ref{tab:qwen2.5_merging_comparison} and \ref{tab:Mistral_results} present the comparison results on large language models. In these tables, the last column, “AVG”, denotes the average performance of the merged model, while the values in parentheses indicate the performance improvement relative to the conference version of CABS. Bold values in each column represent the best performance, whereas underlined values indicate the second-best results. It is evident that CABS+ achieves a substantial performance improvement over the  CABS in the merging experiments on fq2.5-7B and Tsunami-0.5-7B. Its overall performance is highly competitive with the ideal model baseline while further surpassing the recent state-of-the-art method WUDIMerging. In addition, we also conduct experiments under different merging orders, and the results demonstrate that CABS+ exhibits strong robustness to the order of model merging. In the model merging experiments on WildMarcoroni-Variant1-7B and WestSeverus-7B-DPO-v2, the performance of CABS+ is only marginally lower than that of WUDIMerging and remains highly comparable overall. Notably, although the CABS has already surpassed the ideal model baseline, we are still able to achieve further performance improvements on top of it, demonstrating the continued optimization capability of the proposed method.

Beyond the overall comparison, Tables \ref{tab:qwen2.5_merging_comparison} and \ref{tab:Mistral_results} show that pruning strategies generally improve most merging methods on the Qwen2.5 series, but often degrade performance on the Mistral series. In contrast, CABS+ maintains stable performance across both model families, suggesting that its conflict-aware masking mechanism can reduce interference in heterogeneous task settings while better preserving shared representations in homogeneous settings. A detailed analysis is provided in the supplementary material.

\subsection{Results on Small-Scale Language Models}
For the small scale model experiments, we mainly conduct evaluations on GPT-2 and RoBERTa, respectively assessing the performance under different numbers of merged tasks. Specifically, Tables \ref{tab:RoBERTa_Four_results}, \ref{tab:RoBERTa_six_results}, and \ref{tab:RoBERTa_Two_results} present the results on RoBERTa under 4 tasks, 6 tasks, and 2 tasks merging settings, respectively. Tables \ref{tab:gpt2_Two_comparison} and Table \ref{tab:gpt2_six_comparison} correspond to the results on GPT-2 under different task number merging scenarios.

\begin{table*}[!t]
\centering
\caption{Performance of merging six task vectors on RoBERTa models.}
\begin{tabular}{l|c|c|c|c|c|c|c}
\toprule
\textbf{METHOD} & \textbf{RTE} & \textbf{MRPC} & \textbf{CoLA} & \textbf{SST-2} & \textbf{RACE} & \textbf{SQuAD} & \textbf{AVG} \\
\midrule
Finetuned Model & 79.42 & 91.18 & 85.04 & 94.04 & 71.71 & 79.82 & 83.54 \\
\midrule
Task Arithmetic & 67.15 & 79.41 & 72.00 & 85.78 & 56.21 & 38.82 & 66.56 \\
+ Magnitude & \underline{72.56} & 81.13 & \textbf{75.26} & 87.50 & 56.99 & 36.23 & 68.28 \\
+ DARE & 71.12 & 65.44 & 72.48 & 83.37 & 59.57 & 51.39 & 67.23 \\
TIES-Merging & 68.94 & \textbf{86.01} & 66.43 & 83.33 & 40.11 & 47.94 & 65.46 \\
+ DARE & \textbf{74.40} & 83.83 & 72.92 & 56.37 & 60.38 & \underline{53.80} & 66.95 \\
AdaMerging & 49.82 & 73.53 & 70.95 & \textbf{91.06} & 50.75 & \textbf{54.20} & 65.05 \\
+ Magnitude & 57.40 & 69.85 & 70.18 & 87.39 & 59.63 & 47.66 & 65.35 \\
+ DARE & 46.21 & 60.78 & 63.95 & 70.30 & 46.05 & 38.73 & 54.34 \\
WUDIMerging & 47.29 & 68.34 & 69.13 & 51.15 & 49.81 & 35.33 & 53.51 \\
+ Magnitude & 47.21 & 69.37 & 69.11 & 63.42 & \textbf{61.29} & 39.71 & 58.35 \\
+ DARE & 53.65 & 66.47 & 65.36 & 89.45 & 57.71 & 43.04 & 62.61 \\
\midrule
CABS & 68.95 & 82.11 & \underline{73.92} & \underline{90.83} & 58.97 & 42.96 & \underline{69.62} \\
\textbf{CABS+ (ours)} & 71.48 & \underline{84.31} & 70.18 & 90.71 & \underline{60.67} & 45.60 & \textbf{70.49} (+0.87)\\
\bottomrule
\end{tabular}
\label{tab:RoBERTa_six_results}
\end{table*}

\begin{figure}[!t]
\centering
\includegraphics[width=\columnwidth]{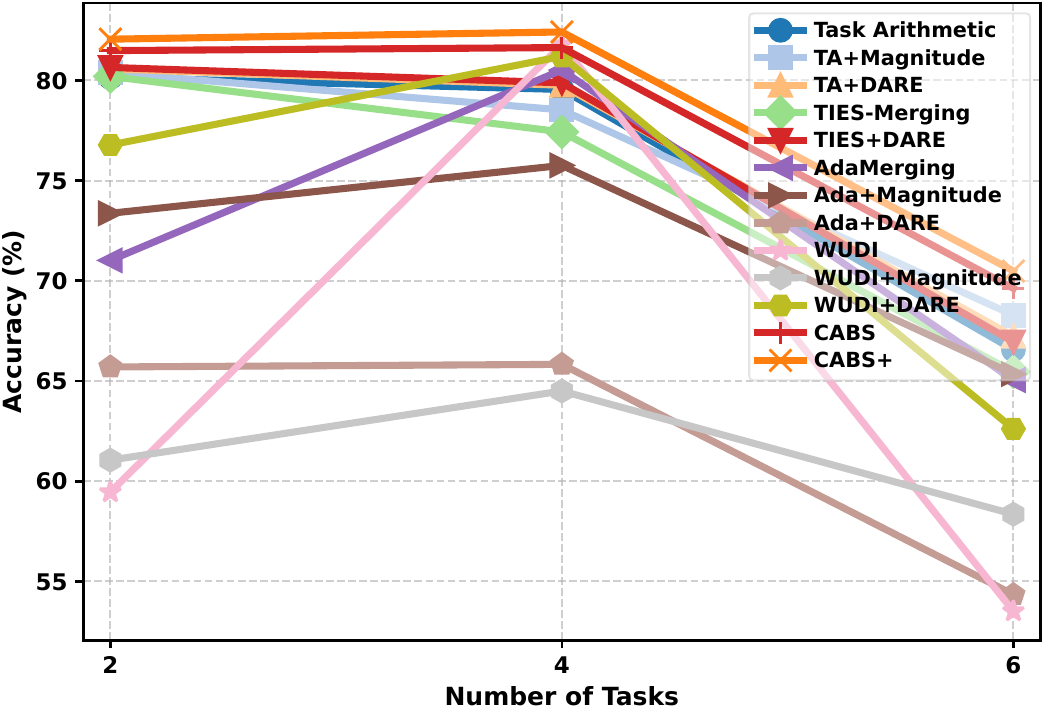}
\caption{Performance comparison of different merging methods on RoBERTa models across varying numbers of tasks.} 
\label{different task_num}
\end{figure}

From Table \ref{tab:RoBERTa_Four_results}, it can be observed that when performing 4 tasks merging on RoBERTa, the current CABS+ consistently achieves significant performance improvements over the original CABS across different task merging orders, while maintaining strong stability under varying merging sequences. At the same time, it clearly outperforms both WUDIMerging and AdaMerging. For the 2 tasks and 6 tasks merging settings reported in Tables \ref{tab:RoBERTa_Two_results} and \ref{tab:RoBERTa_six_results}, respectively, the original CABS itself already surpasses the other compared methods, and CABS+ further improves the performance on top of this strong baseline. To provide a more comprehensive comparison of different methods under varying numbers of merged tasks, we further summarize their average performance on RoBERTa across different task number settings, as shown in Figure \ref{different task_num}. As the number of tasks increases, the overall merging performance declines due to the growing task heterogeneity. This effect becomes particularly pronounced when transitioning from 4 tasks merging to 6 tasks merging, where the inclusion of question answering and multiple choice reading comprehension tasks, specifically SQuAD and RACE, introduces additional complexity. Despite these challenges, CABS+ consistently demonstrates superior performance over the compared methods under different task number configurations, while also exhibiting stronger robustness to variations in the number of merged tasks.

\begin{table}[!t]
\centering
\caption{Performance of merging four task vectors on RoBERTa models.}
\begin{tabular}{l|c|c|c|c|c}
\toprule
\textbf{Method} & \textbf{CoLA} & \textbf{SST-2} & \textbf{RTE} & \textbf{MRPC} & \textbf{Avg} \\
\midrule
Finetuned Model & 85.04 & 94.04 & 79.42 & 91.18 & 87.42 \\
\midrule
Task Arithmetic & 76.32 & 90.83 & 69.68 & 81.37 & 79.55 \\
+ Magnitude & \underline{82.07} & 87.04 & 65.34 & 79.66 & 78.53 \\
+ DARE & 76.99 & 90.14 & 70.76 & 81.13 & 79.76 \\
TIES-Merging & \textbf{82.36} & 86.93 & 61.01 & 79.41 & 77.43 \\
+ DARE & 77.66 & 90.94 & 69.31 & 81.62 & 79.88 \\
AdaMerging & 79.51 & 91.97 & 71.29 & 79.31 & 80.52 \\
+ Magnitude & 69.76 & \textbf{92.55} & 59.99 & 80.73 & 75.76 \\
+ DARE & 40.84 & 85.21 & 52.71 & \textbf{84.56} & 65.83 \\
WUDIMerging & 78.24 & 92.09 & 74.73 & 82.11 & 81.79 \\
+ Magnitude & 69.51 & 71.33 & 47.29 & 69.85 & 64.50 \\
+ DARE & 79.77 & 91.97 & 72.92 & 80.15 & 81.20 \\
\midrule
CABS (CSRM) & 78.24 & \underline{92.32} & 74.37 & 81.62 & 81.64 \\
\textbf{CABS+ (ours)} & 80.88 & 91.55 & 76.62 & 80.64 & 82.42 (+0.78) \\
CABS (MRSC) & 76.89 & 92.09 & 74.73 & \underline{83.09} & 81.70 \\
\textbf{CABS+ (ours)} & 79.48 & 91.32 & \underline{76.99} & 82.07 & \textbf{82.47} (+0.77) \\
CABS (RCMS) & 77.76 & 92.09 & 75.09 & 81.62 & 81.64 \\
\textbf{CABS+ (ours)} & 80.38 & 91.34 & \textbf{77.31} & 80.72 & \underline{82.44} (+0.80) \\
CABS (SCMR) & 78.52 & 91.97 & 73.65 & 82.60 & 81.69 \\
\textbf{CABS+ (ours)} & 81.17 & 91.26 & 75.83 & 81.34 & 82.40 (+0.71) \\
\bottomrule
\end{tabular}
\label{tab:RoBERTa_Four_results}
\end{table}

When performing task merging on GPT-2, we obtain observations similar to those on RoBERTa. As shown in Tables \ref{tab:gpt2_six_comparison} and \ref{tab:gpt2_Two_comparison}, under the 6 tasks merging setting, CABS+ achieves a clear improvement over CABS, with the average performance increasing by 1.56\%, while also surpassing WUDIMerging.
Under the 2 tasks merging setting, even though the original CABS already significantly outperforms the other compared methods and achieves results comparable to the finetuned model baseline, incorporating the AWA strategy into CABS+ still brings performance gain. This further demonstrates the effectiveness of the proposed AWA strategy.

Beyond the aggregate performance trends, AdaMerging and WUDIMerging show noticeable instability across different backbone architectures and task-number settings. In comparison, CABS+ exhibits more stable performance under these varying conditions, which can be attributed to the conflict reduction introduced by CABS pruning and the balanced coefficient optimization enabled by AWA. Detailed analyses of these baseline behaviors and the stability mechanism of CABS+ are provided in the supplementary material.

\begin{table}[!t]
\centering
\caption{Performance of merging RTE-MRPC task pair on RoBERTa models.}
\begin{tabular}{l|c|c|c}
\toprule
\textbf{Method} & \textbf{RTE} & \textbf{MRPC} & \textbf{AVG} \\
\toprule
Finetuned Model & 79.42 & 91.18 & 85.30 \\
\midrule
Task Arithmetic & 73.29 & 87.01 & 80.15 \\
+ Magnitude & \textbf{74.73} & 86.03 & 80.38 \\
+ DARE & 72.92 & 88.24 & 80.58 \\
TIES-Merging & \underline{74.37} & 86.03 & 80.20 \\
+ DARE & 72.56 & 88.73 & 80.65 \\
AdaMerging & 53.07 & 88.97 & 71.02 \\
+ Magnitude & 60.65 & 86.09 & 73.37 \\
+ DARE & 52.71 & 78.68 & 65.70 \\
WUDIMerging & 47.26 & 71.63 & 59.45 \\
+ Magnitude & 47.29 & 74.81 & 61.05 \\
+ DARE & 72.92 & 80.64 & 76.78 \\
\midrule
CABS & 74.01 & \underline{88.97} & \underline{81.49} \\
\textbf{CABS+ (ours)} & 73.20 & \textbf{90.91} & \textbf{82.06} \\
\bottomrule
\end{tabular}
\label{tab:RoBERTa_Two_results}
\end{table}

\begin{table}[!t]
\centering
\caption{Performance of merging CoLA-MRPC task pair on GPT-2 models.}
\begin{tabular}{l|c|c|c}
\toprule
\textbf{Method} & \textbf{CoLA} & \textbf{MRPC} & \textbf{AVG} \\
\midrule
Finetuned Model & 76.80 & 80.39 & 78.60 \\
\midrule
Task Arithmetic & 75.55 & 77.45 & 76.50 \\
+ Magnitude & 76.61 & 79.66 & 78.13 \\
+ DARE & 76.70 & 77.21 & 76.95 \\
TIES-Merging & \underline{76.89} & 77.94 & 77.42 \\
+ DARE & \textbf{77.09} & 76.72 & 76.91 \\
AdaMerging & 69.13 & 72.32 & 70.73 \\
+ Magnitude & 74.50 & \textbf{81.10} & 77.80 \\
+ DARE & 61.35 & 76.78 & 69.07 \\
WUDIMerging & 47.77 & 73.79 & 60.78 \\
+ Magnitude & 47.79 & 76.97 & 62.38 \\
+ DARE & 47.71 & 73.54 & 60.63 \\
\midrule
CABS & 76.41 & 80.88 & \underline{78.65} \\
\textbf{CABS+ (Ours)} & 76.70 & \underline{80.92} & \textbf{78.81} \\
\bottomrule
\end{tabular}
\label{tab:gpt2_Two_comparison}
\end{table}

\begin{table*}[!t]
\centering
\caption{Performance of merging six task vectors on GPT-2 models.}
\begin{tabular}{l|c|c|c|c|c|c|c}
\toprule
\textbf{Method} & \textbf{COLA} & \textbf{MNLI} & \textbf{MRPC} & \textbf{QNLI} & \textbf{QQP} & \textbf{RTE} & \textbf{AVG} \\
\midrule
Fintuned Model & 76.80 & 82.08 & 80.39 & 88.27 & 89.64 & 65.34 & 80.42 \\
\midrule
Task-Arithmetic & 68.94 & 65.30 & 70.59 & 65.28 & 81.35 & 46.21 & 66.28 \\
+ Magnitude & 51.68 & 54.33 & 36.52 & 56.69 & 76.56 & 50.18 & 54.33 \\
+ DARE & 68.74 & 64.13 & 70.83 & 65.13 & 80.48 & 45.85 & 65.86 \\
TIES-Merging & 68.84 & \textbf{68.61} & 69.36 & 66.98 & \underline{82.37} & 44.77 & 66.82 \\
+DARE & 68.46 & 64.60 & 70.10 & 64.91 & 80.75 & 46.21 & 65.84 \\
AdaMerging & 30.87 & 35.45 & 31.62 & 49.46 & 36.82 & \underline{52.71} & 39.49 \\
+ Magnitude & 69.03 & 53.96 & 66.42 & \underline{71.00} & \textbf{84.45} & 48.74 & 65.60 \\
+DARE & 37.91 & 42.55 & 39.41 & 55.78 & 61.04 & \textbf{52.79} & 48.25 \\
WUDIMerging & \textbf{69.61} & 66.75 & 68.41 & \textbf{71.74} & 80.26 & 52.35 & \underline{68.19} \\
+ Magnitude & 62.90 & 57.26 & 33.33 & 61.34 & 78.39 & 52.35 & 57.60 \\
+DARE & 69.22 & 61.53 & \textbf{73.04} & 68.20 & 78.21 & 47.65 & 66.31 \\
\midrule
CABS & 69.13 & 67.47 & 69.61 & 66.34 & 79.05 & 51.29 & 67.15 \\
\textbf{CABS+ (ours)} & \underline{69.24} & \underline{68.55} & \underline{71.36} & 69.65 & 80.80 & 52.65 & \textbf{68.71} (+1.56) \\
\bottomrule
\end{tabular}
\label{tab:gpt2_six_comparison}
\end{table*}

\begin{table}[!t]
\centering
\caption{Efficiency comparison of different merging methods on Roberta and Mistral models in terms of runtime and GPU memory usage.}
\begin{tabular}{l|l|c|c|c}
\toprule
\textbf{Model} & \textbf{Method} & \textbf{AVG} & \textbf{Runtime} & \textbf{GPU Memory} \\
\midrule
Roberta & AdaMerging & 80.52 & 12min & 6.69GB \\
 & WUDIMerging & 81.79 & 3min & 5.10GB \\
 & \textbf{CABS+ (ours)} & 82.42 & 3min & 3.77GB \\
\midrule
Mistral & AdaMerging & 75.90 & 1h & 66.70GB \\
 & WUDIMerging & 76.82 & 4h & 12.04GB \\
 & \textbf{CABS+ (ours)} & 76.71 & 1h & 15.95GB \\
\bottomrule
\end{tabular}
\label{tab:efficiency_comparison_merged}
\end{table}

\subsection{Efficiency Analysis}
To further demonstrate the efficiency of CABS+, we compare CABS+ with the baseline methods CABS, AdaMerging, and WUDIMerging in terms of runtime and GPU memory consumption. Specifically, experiments are performed on RoBERTa under 4 task vectors merging setting, and on Mistral under 2 vectors merging setting. The detailed results are reported in Table \ref{tab:efficiency_comparison_merged}.

For AdaMerging, the optimization process is based on gradient descent. According to the chain rule, all task vectors must be retained in GPU memory simultaneously, as they serve as essential nodes in the computational graph required for gradient computation. In addition, backpropagation requires the construction of the full computational graph and the storage of intermediate activations, leading to substantial GPU memory consumption, especially for large scale models. In contrast, CABS+ adopts a zeroth order optimization strategy, which eliminates the need for gradient computation and significantly reduces both computational overhead and GPU memory usage. The implementation is based on the cma library, where the optimization is executed using Numpy based matrix operations on the CPU. As a result, the GPU is only used for forward inference, serving as a loss evaluation module. After computing the loss, the results are transferred back to the CPU, where the internal distribution of the AWA strategy is updated and the next set of coefficients is generated. The merging of task vectors is then performed on the CPU, and the updated parameters are transferred to the GPU layer by layer. Consequently, at any given time, only the base model parameters and the merged parameters of a single layer are retained in GPU memory, which substantially reduces memory consumption. On the other hand, WUDIMerging adopts a layer wise optimization strategy. For each layer, the corresponding parameters of all task vectors are loaded into GPU memory, and gradient based optimization is performed. This process also requires additional computational graph construction and activation storage. Moreover, frequent data transfer between CPU and GPU is required to load and update layer specific parameters for each task vector. This results in significant overhead due to memory access and 
I/O communication. Such overhead becomes particularly pronounced on large models such as Mistral, where the total runtime can reach up to 4 hours. In comparison, CABS+ reduces the merging time to approximately one quarter of that required by WUDIMerging.

It is worth noting that different strategies are adopted in practice for models of different scales. When performing merging on RoBERTa, the model size is relatively small and GPU memory is sufficient. Therefore, following the common practice adopted by most existing merging methods, both the base model parameters and the task vectors are placed on the GPU for computation and optimization, which minimizes data transfer overhead and accelerates the merging process. In contrast, for large scale models such as Mistral, memory constraints necessitate the use of more memory efficient strategies. Under this setting, when using WUDIMerging, and following its original implementation, only the parameters of the current layer for all task vectors involved in optimization, together with the associated computational graph, are stored in GPU memory, while the remaining layers are kept in CPU memory and loaded on demand. For CABS+, only the base model parameters are maintained in GPU memory for forward inference, while all task vectors, the covariance matrix, and the weighted combination operations are stored in main memory or executed on the CPU. As a result, the GPU memory consumption of WUDIMerging scales linearly with the number of task vectors, corresponding to a space complexity of $\mathcal{O}(n)$, whereas the memory usage of CABS+ remains independent of the number of tasks, corresponding to  $\mathcal{O}(1)$. In the case of two vectors merging, WUDIMerging requires storing only two layers parameters in GPU memory, which leads to relatively lower memory usage. However, due to the additional memory required for gradient computation and intermediate activations, its overall GPU memory consumption remains comparable to that of CABS+. As the number of vectors increases, the memory efficiency of CABS+ becomes more evident. For example, under the four task merging setting on RoBERTa, CABS+ exhibits noticeably lower GPU memory consumption than WUDIMerging.

Combined with the results reported in Table \ref{tab:efficiency_comparison_merged}, CABS+ not only achieves superior average performance on both models, but also effectively avoids the memory overhead introduced by gradient based optimization and the time cost caused by frequent I/O communication. Overall, CABS+ achieves a favorable balance between runtime efficiency and memory usage compared with other methods.

\subsection{Cross-modal Experiments and Ablation Studies}
To further evaluate the generalization capability of CABS+, we conduct additional cross-modal experiments on ViT-B/32 across six vision tasks. Experimental results show that CABS+ achieves an average performance of 82.50, compared with 81.08 for CABS and 82.30 for WUDIMerging, yielding a 1.75\% improvement over CABS while maintaining performance competitive with the state-of-the-art method WUDIMerging.

In addition, ablation studies verify the effectiveness of the asymmetric fitness function and boundary constraints in AWA. Experimental results demonstrate that these components improve optimization stability while substantially reducing the merging time compared with CABS.
The complete results and corresponding experimental settings are provided in the supplementary material.


\begin{figure*}[!t]
\centering
\includegraphics[width=\textwidth]{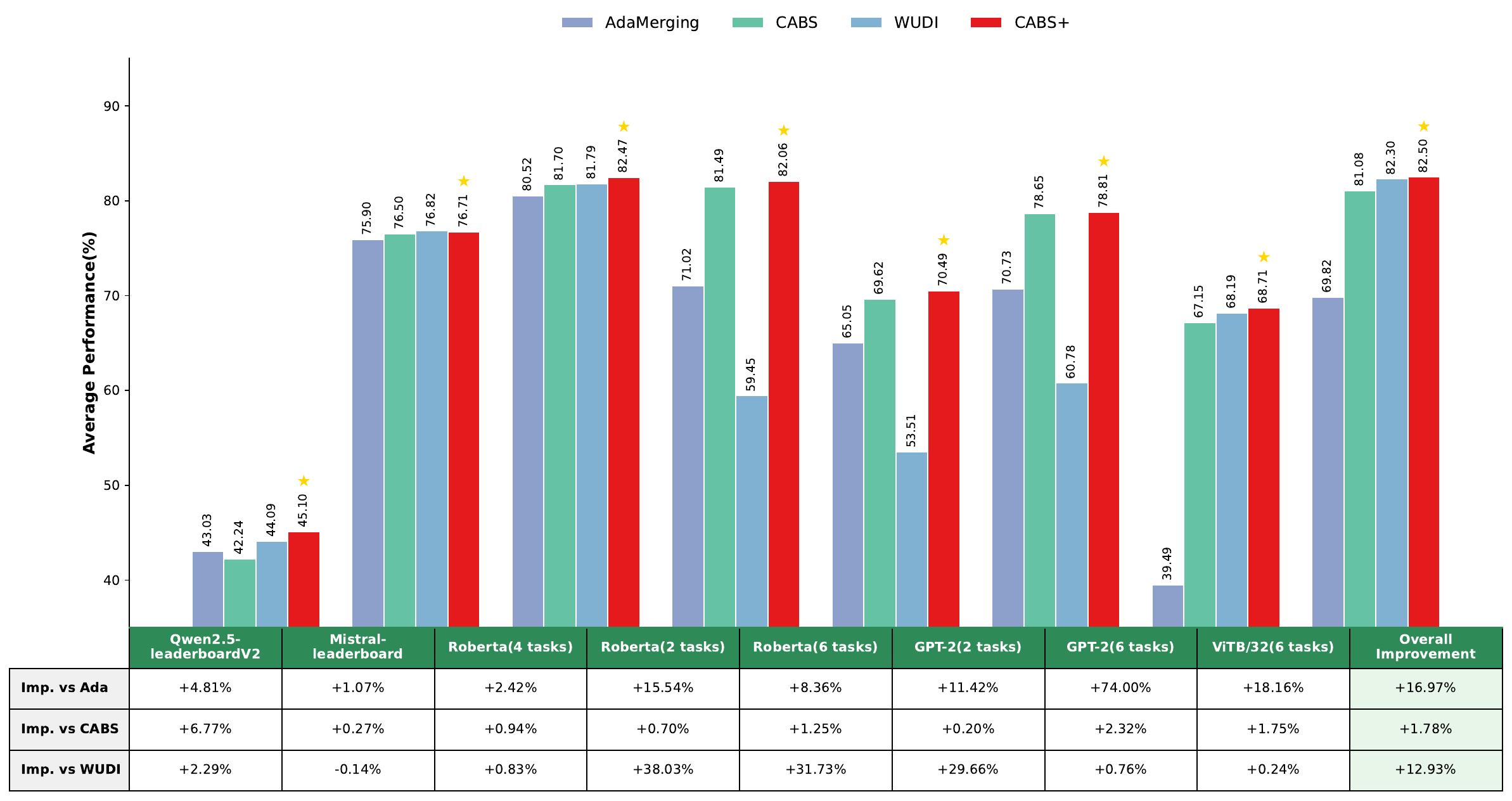}
\caption{Comprehensive Performance and Improvement Comparison Across Methods.}
\label{Overall Impro.}
\end{figure*}

\subsection{Summary}
In summary, CABS+ achieves superior  performance across diverse merging configurations while maintaining low GPU memory consumption, allowing efficient execution on GPUs with limited memory capacity such as V100. 
A comprehensive comparison with AdaMerging, a method of the same category that determines task scaling coefficients without requiring additional training data, and the recent state of the art method WUDIMerging, is presented in Figure \ref{Overall Impro.}.
Due to its robustness across different model architectures and varying numbers of tasks, CABS+ achieves significantly higher average performance, with improvements of 16.97\% and 12.93\% over these two methods, respectively. Furthermore, consistent performance gains are also observed compared with the strong baseline CABS.

\section{Empirical Investigation of Factors Affecting Model Mergeability}
In the previous section, we concentrated on the core question of how to merge and introduced the CABS+ framework as an algorithmic advance designed to improve merging outcomes. However, a logically prior and practically important question that has been largely neglected by existing work is what to merge. This raises a fundamental question: do the intrinsic properties and extrinsic relationships of the source models fundamentally determine the difficulty of merging and the attainable performance ceiling?

Recent work \cite{bowen2024beyond} has begun to focus this question, highlighting that the geometry of task vectors and the degree of inter-model alignment are key determinants of successful model merging. In this section, we present the systematic empirical study of model mergeability. Rather than focusing solely on merging algorithms, we address the following core scientific questions:

\begin{enumerate}[label=Q\arabic*., leftmargin=*, itemsep=0pt]
\item How do a model’s training configuration and optimization history(particularly its key hyperparameters) shape the geometric properties of its task vectors, and consequently influence its mergeability?

\item How do task heterogeneity and finetuning data distribution heterogeneity quantitatively affect the difficulty and effectiveness of model merging?

\item Do architectural differences in the base models and variations in model scale cause finetuned task vectors to exhibit systematically different mergeability characteristics?
\end{enumerate}

By answering these questions, this section seeks not only to uncover the underlying regularities governing model merging, but also to establish a set of principled, scientifically grounded guidelines for model selection. These guidelines are intended to enable researchers and practitioners to make better a priori judgments about the suitability of candidate source models before attempting a merge, thereby shifting model merging from an art of post-hoc hyperparameter fiddling toward a principled practice of pre-merge screening.

To investigate these questions, we design a series of controlled experiments that isolate individual factors, with detailed experimental settings provided in the supplementary material. 
Importantly, to eliminate confounding effects introduced by the merging algorithm itself and to enable an objective assessment of source-model mergeability, all merging experiments in this chapter employ fixed, simple baseline merging methods (primarily non-sparsified Task Arithmetic, with DARE used as a comparison).

\subsection{Impact of Optimization History on Model Mergeability}
To address Q1, we investigate how the optimization history of finetuned models influences their mergeability. Specifically, we focus on two critical optimization hyperparameters, namely the learning rate and the training epochs, and analyze how they shape the geometric properties of task vectors and affect the effectiveness of model merging.

\subsubsection{Effect of Finetuning Learning Rate}\label{LR}

\begin{table}[!t]
\centering
\caption{Effect of Finetuning Learning Rate on Parameter Overlap and Merge Performance}
\begin{tabular}{c|c|c}
\toprule
\textbf{Learning Rate} & \textbf{Parameter Overlap (\%)} & \textbf{Average Accuracy(\%)} \\
\midrule
$1 \times 10^{-6}$ & 85.3 & 79.15 \\
$5 \times 10^{-6}$ & 72.1 & 80.22 \\
$1 \times 10^{-5}$ & 58.6 & 81.05 \\
$2 \times 10^{-5}$ & 45.4 & 81.30 \\
$5 \times 10^{-5}$ & 31.9 & 80.67 \\
\bottomrule
\end{tabular}
\label{tab:learning_rate_overlap}
\end{table}

\begin{table}[!t]
\centering
\caption{Effect of Finetuning Training Epoch on Parameter Overlap and Merge Performance}
\begin{tabular}{c|c|c}
\toprule
\textbf{Training Epochs} & \textbf{Parameter Overlap (\%)} & \textbf{Average Accuracy(\%)} \\
\midrule
1 & 42.5 & 80.11 \\
3 & 46.8 & 81.30 \\
5 & 44.1 & 81.12 \\
10 & 38.2 & 80.45 \\
\bottomrule
\end{tabular}
\label{tab:TrainingEpoch_overlap}
\end{table}

First, we observe a negative correlation between the finetuning learning rate and the parameter overlap ratio. As shown in Table \ref{tab:learning_rate_overlap}, the results reveal a clear and counterintuitive trend: lower learning rates lead to higher parameter overlap. This finding carries important theoretical implications. From the perspective of optimization trajectories, lower learning rates produce smoother and more conservative parameter updates, causing the optimization paths across different tasks to retain and leverage a larger portion of the shared features learned during pretraining. Consequently, task-specific parameter adjustments tend to concentrate within similar subspaces, resulting in higher overlap and greater representational similarity, as measured by metrics such as Centered Kernel Alignment (CKA). In contrast, higher learning rates induce more aggressive and task-specific parameter updates, leading to rapid divergence of optimization trajectories across tasks and a reduced overlap in the parameter space.

Second, the relationship between learning rate and merge performance is non-monotonic. A closer inspection of the data in Table \ref{tab:learning_rate_overlap} indicates that at extremely low learning rates (\(1 \times 10^{-6}\)), corresponding to very high parameter overlap (\(85.3\%\)), the merged model achieves relatively low performance (\(79.15\%\)), consistent with the notion that excessive overlap can induce destructive interference. As the learning rate increases moderately to \(2 \times 10^{-5}\), the overlap decreases and merge performance rises to a peak (\(81.30\%\)). However, when the learning rate becomes too high (\(5 \times 10^{-5}\)), despite continued reduction in overlap, merge performance begins to decline. This phenomenon likely arises because overly high learning rates lead to task-specific overfitting, producing task vectors that are excessively sharp and specialized. While conflicts are reduced, the amount of beneficial shared knowledge across models also diminishes, thereby hindering mergeability.


\subsubsection{Effect of Finetuning Training Epoch}

As shown in Table~\ref{tab:TrainingEpoch_overlap}, increasing the finetuning epoch produces a non-monotonic effect on both parameter overlap and merge performance. The parameter overlap initially increases and reaches the peak at 3 epochs, suggesting that early-stage finetuning (epochs 1–3) primarily reinforces shared features across tasks. In contrast, later-stage finetuning (epochs 5–10) focuses more on learning task-specific features, causing optimization trajectories to diverge and reducing overlap. Merge performance exhibits a typical inverted-U pattern: models with insufficient finetuning (1 epoch) contain incomplete task knowledge, resulting in suboptimal merged accuracy (80.11\%), whereas over-finetuned models (10 epochs) tend to overfit to their respective tasks, producing overly specialized task vectors that are less compatible with other tasks and yielding slightly lower merge performance (80.45\%). In this experiment, the optimal merge performance (81.30\%) occurs at 3 epochs, coinciding with the maximum parameter overlap. These results indicate that finetuning epoch affects mergeability through an optimal finetuning window in which models have sufficiently acquired task knowledge without overfitting, thereby retaining both generalization and compatibility across tasks.

\subsection{Impact of Task and Data Distribution Heterogeneity}
To address Q2, we investigate how heterogeneity arising from task semantics and finetuning data distributions influences model mergeability. Unlike optimization-related factors, which affect mergeability through differences in training dynamics, task and data heterogeneity introduce intrinsic sources of divergence that fundamentally constrain the compatibility of task vectors. When tasks differ substantially in their semantic objectives or data distributions, the corresponding parameter adaptations are more likely to occupy distinct regions of the parameter space, thereby increasing the likelihood of destructive interference during merging.

\subsubsection{Effect of Task Heterogeneity}

To quantify mergeability in a normalized and comparable manner, we adopt the Relative Synergy Score (RSS), defined as:
\begin{equation} 
  \mathrm{RSS}=\frac{\mathrm{Score}_{\mathrm{merged}}-\mathrm{Score}_{\mathrm{ideal}}}{\mathrm{Score}_{\mathrm{ideal}}}\times100\%.
\end{equation}
where $\mathrm{Score}_{\mathrm{merged}}$ denotes the average performance of the merged model across tasks, and $\mathrm{Score}_{\mathrm{ideal}}$ denotes the average performance of the individually finetuned task-specific models. Positive $\mathrm{RSS}$ values indicate synergistic merging, whereas negative $\mathrm{RSS}$ values indicate destructive interference. 


\begin{table}[!t]
\centering
\caption{Impact of Task Heterogeneity on Relative Synergy Score}
\resizebox{\linewidth}{!}{
\begin{tabular}{c|c|c|c|c}
\toprule
\textbf{Level} & \textbf{Task Pair} & \textbf{Score$_{\text{ideal}}$ (\%)} & \textbf{Score$_{\text{merged}}$ (\%)} & \textbf{RSS (\%)} \\
\midrule
Low & MRPC+QQP & 89.50 & 90.15 & +0.73 \\
Medium & SST-2+RTE & 86.21 & 85.95 & -0.30 \\
High & IFEval+GSM8K & 60.39 & 51.90 & -14.06 \\
\bottomrule
\end{tabular}
}
\label{tab:task_heterogeneity_rss}
\end{table}

As shown in Table~\ref{tab:task_heterogeneity_rss}, the experimental results clearly indicate a strong negative correlation between task heterogeneity and synergistic merge performance. In the low-heterogeneity setting, merging produces a positive synergistic gain, with an RSS of \(+0.73\%\). In the medium-heterogeneity setting, the merged model achieves performance close to the ideal baseline but with a slight degradation, yielding an RSS of \(-0.30\%\). In contrast, in the high-heterogeneity scenario, merging causes severe destructive interference, with performance falling well below the ideal model and an RSS of \(-14.06\%\). These results provide quantitative evidence supporting the intuition that merging similar tasks is easier and more effective. 

\subsubsection{Effect of Finetuning Data Distribution}

\begin{table}[!t]
\centering
\caption{Impact of Finetuning Data Distribution Differences on Merge Performance}
\resizebox{\linewidth}{!}{
\begin{tabular}{c|c|c|c|c}
\toprule
\textbf{Level} & \textbf{Task Pair} & \textbf{Score$_{\text{ideal}}$ (\%)} & \textbf{Score$_{\text{merged}}$ (\%)} & \textbf{RSS (\%)} \\
\midrule
Low & IMDb+Yelp & 92.50 & 92.85 & +0.38 \\
High & IMDb+Financial & 90.80 & 84.15 & -7.32 \\
\bottomrule
\end{tabular}
}
\label{tab:data_distribution_rss}
\end{table}

As shown in Table~\ref{tab:data_distribution_rss}, the results reveal a strong positive correlation between data distribution similarity and mergeability. Merging models trained on two highly similar review datasets (IMDb and Yelp) produces a small positive synergistic effect, with $\mathrm{RSS} = +0.38\%$. In contrast, merging models trained on datasets with large domain differences (IMDb and Financial PhraseBank) leads to severe performance degradation, with $\mathrm{RSS} = -7.32\%$. This outcome reflects the fact that differing data distributions guide models to learn distinct features, biases and shortcuts, which may conflict when merged. These findings indicate that data distribution mismatch can substantially reduce mergeability even when the task definitions are identical.


\subsection{Impact of Model Architecture and Scale}
To address Q3, we investigate how differences in base model architecture and model scale influence mergeability. Unlike task and data heterogeneity, which affect mergeability through differences in learned functional representations, architectural and scale differences introduce structural and geometric variations in the parameter space itself. These variations can alter both the alignment and compatibility of task vectors, thereby affecting the feasibility and effectiveness of merging. Understanding these effects is essential for determining whether models derived from different pretrained backbones or with different parameter scales can be reliably merged.

\subsubsection{Effect of Model Architecture}

\begin{table}[!t]
\centering
\caption{Comparison of Average Performance Scores Across Different Model Architectures for Various Sparsification Strategies}
\resizebox{\linewidth}{!}{
\begin{tabular}{l|c|c}
\toprule
\textbf{Method} & \textbf{RoBERTa(Encoder-only)} & \textbf{GPT-2(Decoder-only)} \\
\midrule
Task Arithmetic & 79.55 & 76.50 \\
TA+DARE & 79.76 & 76.95 \\
TA+Magnitude & 78.53 & 78.13 \\
TIES-Merging & 77.43 & 77.42 \\
\midrule
CABS & 81.70 & 78.65 \\
CABS+ & 82.37 & 78.81 \\
\bottomrule
\end{tabular}
}
\label{tab:architecture_comparison}
\end{table}

As shown in Table~\ref{tab:architecture_comparison}, a notable contrast emerges between the two architectures. On RoBERTa (Encoder-only), random pruning (DARE) achieves better performance (79.76) than magnitude pruning (78.53), which is consistent with the previously observed anomalous behavior. However, on GPT-2 (Decoder-only), the trend reverses: magnitude pruning achieves significantly higher performance (78.13) than random pruning (76.95), and its performance approaches that of CABS (78.65).

This observation suggests that the effectiveness of magnitude pruning is not universal, but instead strongly dependent on model architecture. The plausible explanation lies in the differences in information processing mechanisms. Encoder-only models employ bidirectional attention, which may lead to more distributed and cooperative parameter updates, allowing randomly retained parameters to preserve sufficient functionality. In contrast, Decoder-only models rely on autoregressive causal attention, which may result in more localized and concentrated parameter updates, making the preservation of high-magnitude parameters more beneficial for maintaining model performance.

\subsubsection{Effect of Model Scale}

\begin{table}[!t]
\centering
\caption{Impact of Model Scale on Merge Synergy}
\resizebox{\linewidth}{!}{
\begin{tabular}{c|c|c|c}
\toprule
\textbf{Model Scale} & \textbf{Score$_{\text{ideal}}$ (\%)} & \textbf{Score$_{\text{merged}}$ (\%)} & \textbf{RSS (\%)} \\
\midrule
7B & 85.10 & 84.20 & -1.06 \\
13B & 87.50 & 87.45 & -0.06 \\
70B & 90.20 & 91.35 & +1.28 \\
\bottomrule
\end{tabular}
}
\label{tab:model_scale_rss}
\end{table}

As shown in Table~\ref{tab:model_scale_rss}, a clear trend emerges: mergeability improves consistently as model scale increases. On the 7B model, merging results in mild destructive interference, with an RSS of $-1.06$\%. On the 13B model, this interference nearly disappears, yielding an RSS of $-0.06$\%. On the 70B model, merging produces a clear synergistic effect, with an RSS of $+1.28$\%.

This trend can be attributed to two key factors: knowledge decoupling and parameter redundancy. Larger models possess substantially more parameters, enabling them to store task-specific knowledge in more distinct and potentially near-orthogonal parameter subspaces. This implicit knowledge decoupling reduces the likelihood of destructive interference between task vectors during merging. Additionally, large models exhibit higher parameter redundancy. Even when some parameters are negatively affected by merging conflicts, other redundant parameters can compensate for the loss, resulting in improved robustness and overall merge performance. This finding provides a new perspective on the advantages of large-scale models: beyond superior single-task performance, they also offer intrinsic benefits for multi-task knowledge integration through model merging.


\subsection{Summary of Key Findings}

Overall, the empirical results identify six factors that influence model mergeability, including learning rate, finetuning epoch, task heterogeneity, data distribution heterogeneity, model architecture, and model scale. These findings suggest that model merging performance is not determined solely by the merging algorithm, but also by the intrinsic properties and relationships of the source models.
The key findings and contributions are summarized as follows:
\begin{itemize}
\item Training configurations, particularly learning rate and finetuning epoch, are critical intrinsic factors affecting model mergeability. These factors directly influence the geometric structure and compatibility of task vectors, thereby determining the effectiveness of model merging.
\item Task heterogeneity constitutes a fundamental obstacle to successful merging. To enable quantitative analysis of this effect, a new metric, the Relative Synergy Score (RSS), is proposed to measure synergistic gain and destructive interference in a normalized and interpretable manner.
\item Finetuning data distribution is another key factor affecting mergeability. Even when task definitions are identical, differences in training data distributions can significantly reduce model compatibility. This finding highlights the importance of carefully examining training data sources and distributions during the model selection stage prior to merging.
\item Model architecture constitutes an important determinant of mergeability characteristics. Encoder-only and Decoder-only architectures exhibit systematically different responses to sparsification strategies, indicating that merge behavior is strongly architecture-dependent.
\item Model scale positively correlates with mergeability. Larger models exhibit stronger knowledge decoupling and parameter redundancy, enabling more effective integration of task-specific knowledge.
\end{itemize}

These findings deepen the scientific understanding of model merging and provide practical, actionable guidelines for model selection prior to merging. They constitute an independent and complementary contribution alongside the CABS series methodology.

\section{Conclusion}

This paper proposed CABS+, an enhanced model merging framework that extends CABS with the Adaptive Weight Allocation strategy. By replacing grid search with gradient-free coefficient optimization, CABS+ improves merging efficiency, reduces GPU memory consumption, and promotes more balanced performance improvements across tasks.
We further conducted the systematic empirical study of model mergeability and introduced the Relative Synergy Score as the quantitative measure. The analysis identifies key factors affecting mergeability and provides practical guidance for model selection before merging.
Extensive experiments across diverse tasks, model architectures, and model scales demonstrate that CABS+ achieves robust performance with favorable efficiency. Future work will explore multimodal and heterogeneous model merging to further broaden its applicability.

\vfill

\end{document}